\documentclass{article}

\usepackage{microtype}
\usepackage{caption}
\usepackage{subcaption}
\usepackage{booktabs}
\usepackage[
        hidelinks,
        colorlinks=false,
        pdftex,
        pdfpagelabels,
        bookmarks,
        hyperindex,
        hyperfigures,
        linktoc=all,
        bookmarksnumbered=true,
        bookmarksopen=true
    ]{hyperref}
\hypersetup{}

\usepackage[T1]{fontenc}    \usepackage{amsfonts}       \usepackage{amssymb}
\usepackage{amsmath}
\usepackage{nicefrac}
\usepackage[utf8]{inputenc}

\usepackage[pdftex]{graphicx}
\graphicspath{{images/}}
\DeclareGraphicsExtensions{.pdf,.jpeg,.png}

\usepackage{algorithmic}

\usepackage{url}

\usepackage[dvipsnames,table]{xcolor}

\usepackage{wrapfig}

\usepackage{natbib}

\usepackage{tikz}
\usetikzlibrary{bayesnet}

\usepackage{environ}
\makeatletter
\newsavebox{\measure@tikzpicture}
\NewEnviron{scaletikzpicturetowidth}[1]{  \def\tikz@width{#1}  \begin{lrbox}{\measure@tikzpicture}  \BODY
  \end{lrbox}  \pgfmathparse{#1/\wd\measure@tikzpicture}    \BODY
}
\makeatother

\usepackage{xr}

\makeatletter
\newcommand*{\addFileDependency}[1]{  \typeout{(#1)}
  \@addtofilelist{#1}
  \IfFileExists{#1}{}{\typeout{No file #1.}}
}
\makeatother

\newcounter{docpart}
\def\savestatus{  \newwrite\tempfile  \immediate\openout\tempfile=docstatus\arabic{docpart}.dat  \immediate\write\tempfile{\thesection}  \immediate\write\tempfile{\theequation}  \immediate\closeout\tempfile}
\newcounter{olddocpart}

\newcommand{\bs}{\boldsymbol}

\newcommand{\TZK}{{TzK }}

\newcommand{\ki}{{\mathrm{e}}^{i}}
\newcommand{\kj}{{\mathrm{e}}^{j}}
\renewcommand{\k}{\bar{{\mathrm{e}}}}

\newcommand{\Ci}{{\bs{c}^{i}}}
\newcommand{\Cj}{{\bs{c}^{j}}}
\newcommand{\C}{\bar{\bs{c}}}
\newcommand{\fCi}{f_{\Ci}}

\newcommand{\Ki}{{{\bs k}^{i}}}

\newcommand{\K}{{\bar{{\bs k}}}}

\newcommand{\T}{{\bs t}}
\newcommand{\fT}{f_{\T}}
\newcommand{\Z}{{\bs z}}

\newcommand{\Y}{{\bar{\bs y}}}

\newcommand{\params}{{\bs \theta}}
\newcommand{\decparams}{{\bs \psi}}

\newcommand{\encparams}{{\bs \phi}}

\newcommand{\M}{\mathcal{M}}
\newcommand{\Mopt}{\M_{\params}}

\newcommand{\Mmodel}{\hat{\M}_{\params}}

\newcommand{\Mdec}{p^{dec}_{\decparams}}
\newcommand{\Menc}{p^{enc}_{\encparams}}

\newcommand{\pjoint}{\mathcal{P}}
\newcommand{\penc}{p^{enc}}
\newcommand{\pdec}{p^{dec}}

\newcommand{\Tdim}{D}
\newcommand{\Cdim}{C}

\newcommand{\Knum}{K}

\newcommand{\DKL}[2]{\mathcal{D}_\text{KL}\left(#1\|\, #2\right)}

\newcommand{\E}[2]{\mathbb{E}_{#1}\left[#2\right]}

\newcommand{\DETDF}[2]{ | \det \frac{\partial {#1}}{\partial {#2}} | }

\newcommand{\eg}{{\em e.g.}}
\newcommand{\ie}{{\em i.e.}}

\usepackage[accepted]{icml2019}

\icmltitlerunning{TzK: Flow-Based Conditional Generative Model}

\begin{document}

\twocolumn[
\icmltitle{TzK: Flow-Based Conditional Generative Model}

\icmlsetsymbol{equal}{*}

\begin{icmlauthorlist}
\icmlauthor{Micha Livne}{to,vec}
\icmlauthor{David Fleet}{to,vec}
\end{icmlauthorlist}

\icmlaffiliation{to}{Department of Computer Science, University of Toronto}
\icmlaffiliation{vec}{Vector Institute, Toronto}

\icmlcorrespondingauthor{Micha Livne}{mlivne@cs.toronto.edu}
\icmlcorrespondingauthor{David Fleet}{fleet@cs.toronto.edu}

\icmlkeywords{Machine Learning, ICML}

\vskip 0.3in
]

\printAffiliationsAndNotice{}
\begin{abstract}
We formulate a new class of conditional generative models based on probability flows.
Trained with maximum likelihood, it provides efficient inference and sampling
from class-conditionals or the joint distribution, and does not require
{\em a priori} knowledge of the number of classes or the relationships between classes.
This allows one to train generative models from multiple,
heterogeneous datasets, while retaining strong prior models over
subsets of the data (e.g., from a single dataset, class label, or attribute).
In this paper, in addition to end-to-end learning, we show how one can learn a single model from
multiple datasets with a relatively weak Glow architecture, and then extend it by conditioning on different
knowledge types (\eg, a single dataset).
This yields log likelihood comparable to state-of-the-art, compelling samples from conditional  priors.
\end{abstract}

\section{Introduction} \label{sec:introduction}

The goal of representation learning is to learn structured, meaningful
latent representations given large-scale unlabelled datasets.
It is widely assumed that such unsupervised learning will support myriad
downstream tasks, some of which may not be known {\em a priori}  \cite{Bengio2013}.
To that end it is useful to be able to train on large amounts of hetergeneous
data, but then use conditional priors that isolate specific sub-spaces or manifolds
from the broader data distribution over the observation domain.

Building on probability flows \cite{Dinh2014,Dinh2016a,Kingma2018},
this paper introduces a flexible form of conditional generative model.
It is compositional in nature, without requiring {\em a priori} knowledge
of the number of classes or the relationships between classes.
Trained with maximum likelihood, the framework allows one to learn
from heterogeneous datasets in an unsupervised fashion, with concurrent
or subsequent specialization to sub-spaces or manifolds of the observation
domain, \eg, conditioning on class labels or attributes.
The resulting model thereby supports myriad downstream tasks, while providing
efficient inference and sampling from the joint or conditional priors.

\subsection{Background}
\label{sec:background}

There has been significant interest in learning generative models in recent years.
Prominent models include variational auto-encoders (VAE), which maximize a variational lower
bound on the data log likelihood \cite{Rezende2014,Kingma2013,Berg2018,Papamakarios2017,Kingma2016},
and generative adversarial networks (GAN), which use an adversarial discriminator to enforce a
non-parametric data distribution on a parametric decoder or encoder
\cite{NIPS2014_5423,Makhzani2018,Makhzani2015,Chen2016}.
Inference, however, remains challenging for VAEs and GANs as neither model includes a
probability density estimator \cite{schmah2009,Papamakarios2017,Dinh2016a,Dinh2014}.

Auto-regressive models \cite{Germain2015,bengio1999,Larochelle2011} and normalizing
flows \cite{Dinh2014,Dinh2016a,Rezende2015,Kingma2018} train with maximum likelihood (ML),
avoiding approximations by choosing a tractable parameterization of probability density.
Auto-regressive models assume a conditional factorization of the density function,
yielding a tractable joint probability model.
Normalizing flows represent the joint distribution with a series of invertible
transformations of a known base distribution, but are somewhat problematic
in terms of the memory and computational costs associated with large volumes
of high-dimensional data (\eg images).
While invertibility can be used to trade memory with compute requirements
\cite{Chen2018,Gomez2017}, training powerful density estimators remains challenging.

The attraction of unsupervised learning stems from a desire to exploit vast
amounts of data, especially when downstream tasks are either unknown {\em a priori},
or when one lacks ample task-specific training data.
And while samples from models trained on heterogeneous data may
not resemble one's task domain per se, conditional models can be used to isolate
manifolds or sub-spaces associated with particular classes or attributes.
The \TZK framework incorporates task-specific conditioning in a flexible manner.
It supports end-to-end training of the full model.  Or one to train a
powerful density estimator once, retaining the ability to later extend it
to new domains, or specialize it to sub-domains of interest.
We get the advantages of large heterogenous datasets, while retaining
fidelity of such specialized conditional models.

Existing conditional generative models allow one to sample from
sub-domains of interest (\eg, \cite{Makhzani2018,Chen2016,Dupont2018}),
but they often require that the structure of the data and latent representation
be known {\em a priori} and embedded in the network architecture.
For example, \cite{Chen2016,Makhzani2018} allow unsupervised
learning but assume the number of (disjoint) categories is given.
In doing so they fix the structure of the latent representation to include a
1-hot vector over categories at the time of training.
Such models are therefore re-trained from scratch if labels change, or if
new labels are added, \eg by augmenting the training data.

\citet{Kingma2018} train a conditional prior post hoc, given an existing Glow model.
This allows them to condition an existing model on semantic attributes,
but lacks the corresponding inference mechanism.
A complementary formulation, augmenting a generative model with a post hoc discriminator,
is shown in \cite{Oliver2018}.

Inspired by \cite{Kingma2018,Oliver2018}, \TZK incorporates
conditional models with discriminators and generators, all trained jointly.
The proposed framework can be trained unsupervised on large volumes of data,
yielding a generic representation of the observation domain (\eg, images),
while explicitly
supporting the semi-supervised learning of new classes in an online fashion.
Such conditional models are formulated to be compositional, without a prior knowledge
of all classes, and exploiting similarity among classes with a joint latent representation.

Finally, the formulation below exhibits an interesting connection between
the use of  mutual information (MI) and ML in representation learning.
The use of MI is prevalent in learning latent representations
\cite{Hjelm2018,Chen2016,Dupont2018,Klys2018}, as it provides a measure of
the dependence between random variables. Unfortunately, MI is hard to compute;
it is typically approximated or estimated with non-parametric approaches.
A detailed analysis is presented in \cite{Hjelm2018},
which offers scalability with data dimensionality and sample size.
While it is intuitive and easy to justify the use of MI to enforce a relationship
between random variables  (\eg, dependency \cite{Chen2016} or independence \cite{Klys2018}),
MI is often used as a regularizer to extend an existing model.
The \TZK formulation offers another perspective, showing how MI arises naturally
with the ML objective, following the assumption that a target distribution can be
factored into (equally plausible) encoder and decoder models.  We exploit a lower bound
that allows indirect optimization of MI, without estimating MI directly.

\textbf{Contributions:}
We introduce a conditional generative model based on probability density
normalizing flows, which is flexible and extendable.
It does not require that the number of classes be known {\em a priori}, or that classes are
mutually exclusive.
One can train a powerful generative model on unlabeled samples
from multiple datasets, and then adapt the structure of the latent
representation as a function of specific types of knowledge in an online fashion.
The proposed model allows
high parallelism when training multiple tasks while maintaining a joint distribution over
the latent representation and observations, all with efficient inference and sampling.

\section{\TZK Framework}
\label{sec:tzk}

We model a joint distribution over an observation domain (\eg, images)
and latent codes (\eg, attributes or class labels).
Let observation $\T \in \mathbb{R}^{\Tdim}$ be a random variable associated through
a probability flow with a latent state variable $\Z \in \mathbb{R}^{\Tdim}$  \cite{Dinh2014,Dinh2016a,Rezende2015,Kingma2018}.
In particular, $\Z$ is mapped to $\T$ through a smooth
invertible mapping $\fT:\mathbb{R}^{\Tdim} \rightarrow \mathbb{R}^{\Tdim}$, \ie, $\T=\fT(\Z)\label{eq:T-flow}$.
As such, $\fT$ transforms a base distribution $p(\Z)$ (\eg, Normal) to
a distribution $p(\T)=p(\Z) \,\DETDF{\fT}{\Z}^{-1}$ over the
observation domain.
Normalizing flows can be formulated as a joint distribution
$p(\Z,\T)=\delta (\Z - \fT^{-1} (\T )) p(\T)=\delta (\T - \fT (\Z) ) p(\Z)$,
but for notational simplicity we can omit $\T$ or $\Z$ from
probability distributions by trivial marginalization of one or the other.

\begin{figure}[t]
    \centering
    \begin{minipage}[b]{.5\columnwidth}
    \centering \scalebox{0.85}{\begin{tikzpicture}

    \node[obs]                               (T) {$\T$};
  \node[latent, above=of T]                (Z) {$\Z$};
  \node[latent, above=of Z, xshift=1.2cm] (ki) {$\ki$};
  \node[obs, above=of Z, xshift=-1.2cm]     (C) {$\Ci$};

  \edge{Z} {T} ; \edge {T} {ki,C} ; \edge {ki} {C} ;
\plate {Ki} {(C)(ki)} {$\Knum$} ;

\end{tikzpicture}
}
    \subcaption{encoder}\label{fig:tzk-model-encoder}
    \end{minipage}%
   \begin{minipage}[b]{.5\columnwidth}
    \centering \scalebox{0.85}{\begin{tikzpicture}

    \node[obs]                               (T) {$\T$};
  \node[latent, above=of T]                (Z) {$\Z$};
  \node[latent, above=of Z, xshift=1.2cm] (ki) {$\ki$};
  \node[obs, above=of Z, xshift=-1.2cm]     (C) {$\Ci$};

  \edge{Z} {T} ; \edge {ki,C} {Z} ; \edge {C} {ki} ;
\plate {Ki} {(C)(ki)} {$\Knum$} ;

\end{tikzpicture}
}
    \subcaption{decoder}\label{fig:tzk-model-decoder}
    \end{minipage}
    \vspace*{-0.6cm}
    \caption{\TZK framework models $\pjoint(\T,\K)$, a joint distribution over task domain $\T$ and
     multiple latent codes $\K = \{\Ci, \ki \}_{i=1}^{\Knum}$ with a dual encoder/decoder.
    The framework offers explicit representation of sub-domains of interest in $\pjoint(\T,\K)$ by
    conditioning on the latent codes which comprise a single compositional model.
    }
    \label{fig:tzk-model}
\end{figure}
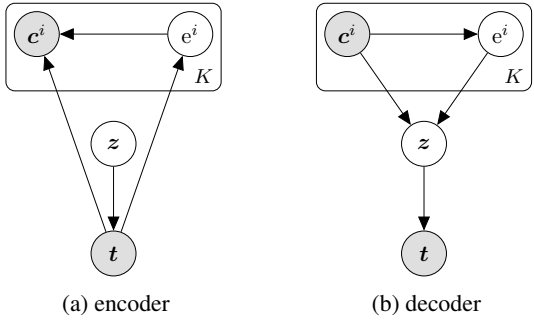

For conditional generative models within the \TZK framework, the latent state
$\Z$ is conditioned on a latent code (see Fig.\ \ref{fig:tzk-model}b).
As such, they capture distributions within the observation domain
associated with subsets of the training data, or subsequent labelled data.
To this end, let $\Ki$ be a hybrid discrete/continuous random variable
$\Ki \equiv ( \ki,\Ci )$, where $\ki \in \{0,1 \}$
and $\Ci \in \mathbb{R}^{\Cdim}$, similar to \cite{Chen2016,Dupont2018}.
We refer to $\Ki$ as \textit{knowledge} of type $i$, while $\Ci$ is the latent
code of knowledge $i$, a structured latent representation of $\T$.
We call $\ki$ the existence of knowledge $i$, a binary variable
that serves to indicate whether or not $\T$ can be generated by $\Ci$.

To handle multiple types of knowledge, let $\K = \{  \Ki \}_{k=1}^{\Knum}$
denote the set of latent codes associated with $\Knum$ knowledge types.
Importantly, we do not assume that knowledge
types correspond to mutually exclusive class labels.
Rather, we allow varying levels of interaction between knowledge classes under
the \TZK framework. This avoids the assumption of mutually exclusive classes
and allows a \TZK model to share a learned representation between
similar classes, while still being able to represent distinct classes.

\subsection{Formulation} \label{sec:tzk-formulation}

Our goal is to learn a probability density estimator of the joint
distribution $\pjoint (\T, \K )$. In terms of an encoder-decoder,
for effective inference and sample generation, we model $\pjoint$
in terms of two factorizations, i.e.,
\begin{eqnarray}
\penc\left(\T,\K\right) &=& p\left(\K|\T\right)\, p\left(\T\right)
\label{eq:tzk-encoder} \\
\pdec\left(\T,\K\right) &=&  p\left(\T|\K\right)\, p\left(\K\right)
\label{eq:tzk-decoder}
\end{eqnarray}
The encoder factorization in Eq.\ \eqref{eq:tzk-encoder} makes $p(\K|\T)$
explicit, which is used for inference of the latent  code given $\T$.
The decoder in Eq.\ \eqref{eq:tzk-decoder} makes $p(\T|\K)$ explicit for
generation of samples of $\T$ given a latent code $\K$.

As noted by \cite{Kingma2013,Agakov2004,Rezende2015,Chen2016},
inference with the general form of the posterior $p( \K | \T )$ is
challenging. Common approaches resort to variational approximations
\cite{Kingma2013,Rezende2015,Kingma2016}.
A common relaxation in the case of discrete latent codes is the assumption of
independence (\eg, $p(\k|\T,\C )=\prod_i p(\ki|\T,\C )$ ).
Alternatively, one can assume that such binary codes represent
mutually exclusive classes, e.g., with a single categorical random variable.
But this makes it difficult to model attributes, for which the presence or
absence of one attribute may be independent of other attributes,
or to allow for the fact that one image may belong to two different
classes (\eg, it might be present in more than one database).

Here we design \TZK to avoid the need for mutual exclusivity, or the
need to specify the number of classes a priori, instead allowing the model
to be extended with new classes, and to learn and exploit some degree of
similarity between classes.
To that end we assume that knowledge types exhibit statistical independence,
expressed in terms of the following encoder factorization,
\begin{eqnarray}
\penc\left(\T,\K\right) ~=~  p\left(\T\right)\prod_{i}p\left(\Ki|\T\right) ~,
\label{eq:tzk-encoder-compositionality}
\end{eqnarray}
and the corresponding decoder factorization
\begin{eqnarray}
\pdec(\T,\K)
&=& p(\K)\, \frac{p(\K|\T)\,p(\T)}{p(\K)} \nonumber \\
&=& p(\K)\,p(\T)\prod_{i}\frac{p(\Ki|\T)\,p(\T)}{p(\Ki)\,p(\T)} \nonumber \\
&=&  \frac{\prod_{i}p\left(\T|\Ki\right)p\left(\Ki\right)}{p\left(\T\right)^{\Knum-1}} \label{eq:tzk-decoder-compositionality}
\end{eqnarray}
It is by virtue of this particular factorization that a \TZK model is easily
extendable with different knowledge types (and conditional models) in an online fashion.

Taking the hybrid form of knowledge codes into account,
as in Fig.\ \ref{fig:tzk-model}, the model is further factored as follows:
\begin{eqnarray}
p\left(\Ki|\T\right)&=&p\left(\Ci|\ki, \T\right)p\left(\ki|\T\right)
\label{eq:tzk-encoder-KT} \\
p\left(\T|\Ki\right)&=&p\left(\T|\ki, \Ci\right)
\label{eq:tzk-decoder-TK} \\
p\left(\Ki\right)&=&p\left(\ki|\Ci\right)p\left(\Ci\right)
\label{eq:tzk-decoder-K} ~.
\end{eqnarray}
Here, $p\left(\ki=1|\T\right)$ and $p\left(\ki=1|\Ci\right)$ act as
discriminators for binary variable $\ki$, conditioned on $\T$ and $\Ci$ respectively.

Finally, the factors of the encoder and decoder in  \eqref{eq:tzk-encoder-compositionality}
- \eqref{eq:tzk-decoder-K} are parametrized in terms of neural networks.
Accordingly, denoting the parameters of the encoder and decoder by $\encparams$ and
$\decparams$, in what follows we write the parametrized model encoder and decoder
as  $\Menc ( \T, \K )$ and $\Mdec ( \T, \K )$.
(In what follows we use this more concise notation for the encoder and decoder,
except where we need the explicit factorization in terms of $\Ki$, $\Ci$ and $\ki$.)
Details of our implementation are described in Sec.\ \ref{sec:implementation}.

\subsection{Learning} \label{sec:tzk-learning}

We would like to train a parametric model of the joint distribution
$\pjoint (\T,\K )$ with the dual encoder/decoder factorization defined in
Eqs.\ \eqref{eq:tzk-encoder-compositionality} - \eqref{eq:tzk-decoder-K}.
Following the success of \cite{Dinh2014,Dinh2016a,Kingma2018} with high-dimensional
distributions, we opt to estimate the model parameters using maximum likelihood.

We aim to learn a single probabilistic model of $\pjoint (\T,\K )$,
comprising a consistent encoder-decoder, with the factorization given
above, and a shared flow $\fT$.
To do so, we define a joint distribution $\Mopt\left(\T,\K\right)$ with parameters
$\params = \{ \encparams, \decparams \} $. Expressed as a linear mixture, $\Mopt$ is
randomly selected to be $\Menc$ or $\Mdec$ with equal probability, \ie,
\begin{equation}
\Mopt\left(\T,\K\right) =
\frac{1}{2}\left( \,\Menc\left(\T,\K\right) + \Mdec\left(\T,\K\right)\, \right) ~.
\label{eq:tzk-encoder-decoder}
\end{equation}
Choosing the mixing coefficients to be equal reflects our assumption of a
dual encoder/decoder parametrization of the same underlying joint distribution.
There are other ways to combine $\penc$ and $\pdec$ into a single model;
we chose this particular formulation because it yields a very effective learning algorithm.

Learning maximum likelihood parameters entails maximizing
$\E{\K,\T\sim \pjoint}{\log\Mopt (\K,\T)}$ with respect to $\params$; equivalently,
\begin{eqnarray}
\params^{*} =  \arg\max_{\params} -H(\pjoint, \Mopt ) ~,
\label{eq:tzk-entropy-loss}
\end{eqnarray}
where $H\left( \cdot, \cdot \right)$ is the usual cross-entropy.
Instead of optimizing the negative cross entropy directly, which
can be numerically challenging, here we advocate the optimization
of a lower bound on $-H(\pjoint, \Mopt )$.
Using Jensen's inequality it is straightforward to show that
$\log (\frac{1}{2} \Menc + \frac{1}{2} \Mdec ) \ge \frac{1}{2} \log\Menc + \frac{1}{2} \log\Mdec$,
and as a consequence,
\begin{eqnarray}
~ -H(\pjoint, \Mopt ) \,\ge\,
-\frac{1}{2} \left[ \, H(\pjoint,\Menc) + H(\pjoint,\Mdec) \, \right] \, .
\label{eq:tzk-entropy-loss-lower-bound}
\end{eqnarray}

The lower bound turns out to be very useful because, among other
things, it encourages consistency between the encoder and decoder.
To see this, we examine the bound in greater detail.
With some algebraic manipulation, ignoring expectation in
Eq.\ (\ref{eq:tzk-entropy-loss-lower-bound}), one can derive
the following:
\begin{equation}
 \frac{\log \Menc\! \!+\! \log \Mdec}{2} =
\log\Mopt -
\log\!\frac{
\sqrt{\frac{\Menc}{\Mdec}} \! +\!
\sqrt{\frac{\Mdec}{\Menc}}
}{2}  \, . \!\!\!
\label{eq:tzk-entropy-loss-regularized}
\end{equation}
This implies that maximization of the lower bound (the expectation of the LHS)
entails maximization of the expectation of the two terms on the RHS, the first
of which is $-H(\pjoint, \Mopt )$.
The expectation of the second term on the RHS of Eq.\ (\ref{eq:tzk-entropy-loss-regularized})
can be viewed as a regularizer that encourages the encoder and  decoder to assign
similar probability density to each datapoint.
Importantly, it obtains its upper bound of zero when $\Menc = \Mdec$, in which case
the inequality in Eq.\ (\ref{eq:tzk-entropy-loss-lower-bound}) becomes equality.
In practice, we find the bound is tight.

It is also interesting to note that $-H(\pjoint, \Mopt )$ itself is a lower
bound on $-H\left(\pjoint\right)$, since $-H\left(p\right) \ge -H\left(p, q\right)$
for any distributions $p$ and $q$.
If  $\pjoint (\K,\T )$ satisfies the factorization
of the \TZK model in Eqs.\ (\ref{eq:tzk-encoder-compositionality}) - (\ref{eq:tzk-decoder-K})
then the entropy of the joint distribution can be expressed as
\begin{eqnarray}
-H(\K,\T) &=&
-H(\T)\, -\, \sum_{i}H(\Ki) \,  \nonumber \\
& \ &  +\, \frac{1}{2}\sum_{i}\left[\,I(\Ki;\T) + I(\Z;\Ki)\, \right] \, ,~
\label{eq:tzk-entropy}
\end{eqnarray}
where $H(\T)$ and $H(\K)$ denote entropy of marginal distributions,
and $I(\K,\T)$ and $I(\Z;\Ki)$ denote mutual information, for which
all expectations are with respect to $\K,\T \sim \pjoint$.
(The derivation of Eq.\ \eqref{eq:tzk-entropy} is given in the supplemental material.)
Eq.\ \eqref{eq:tzk-entropy} suggests that maximizing the MI between observations
and latent codes here follows from a design choice, for a model that can equally
well "understand" (encode) and "express" (decode) an independent set of latent
codes (as in Eqs.\ \eqref{eq:tzk-encoder-compositionality} and
\eqref{eq:tzk-decoder-compositionality}), within a shared observation domain.

We claim that the assumption of independent latent codes is a relatively mild
assumption, and has little affect on the ability of the model to represent
$p (\T | \Y )$ for a random variable $\Y $ over the same domain as $\K$.
A sufficiently expressive flow $\T = \fT (\Z )$ will
allow for $p (\T|\Y) = \pjoint ( \fT(\Z )|\K )$ for arbitrary $\Y \sim p(\Y)$, and
$p(\K) = \prod_i p(\Ki)\,$ \cite{Dinh2014}.
Effectively, we approximate the relationship between factors of $\Y$ by learning the relation
between conditional distribution of independent factors over the same observation domain.
Although such an approximation may not exist for priors, it is effective when dealing with
conditional distributions.  As we demonstrate below, \TZK can learn meaningful representations
of the joint knowledge $p(\Y)$.

\section{Implementation}
\label{sec:implementation}

The \TZK model comprises probability distributions defined in
Eqs.\ (\ref{eq:tzk-encoder-compositionality}) - (\ref{eq:tzk-decoder-K}).
Each can be treated as a black box object with the functionality of a
probability density estimator, returning $\log p\left( x \right)$ given $x$,
and a sampler, returning $x \sim p\left(x\right)$ given $p\left(x\right)$.
The specific implementation choices outlined were made for the ease and efficiency of training.

We adopt a Glow-based architecture for probability density estimators, using
reparametrization \cite{Bernardo2003,Williams1992} and back-propagation with
Monte Carlo \cite{Rezende2014} for efficient gradient-based optimization
(\eg, see \cite{Rezende2015}).
Our flow architecture used fixed shuffle permutation rather than invertible $1\!\times\!1$
convolution used in \cite{Kingma2018} as we found it to suffer from accumulated numerical error.
We implemented \TZK in Pytorch, using $swish\left(x\right) = x \cdot \sigma\left(x\right)$
non-linearity \cite{Ramachandran2017} instead of ReLU as the activation function.
We found that the ReLU-based implementation converged more slowly because of
the truncation of gradients for negative values.

We implemented separated $p\left(\Ci|\ki, \T\right)$ and $p\left(\T|\ki, \Ci\right)$ for $\ki \in {0,1}$
with regressors from $\T$ and $\Ci$ to parameters of distributions over $\Ci$ and $\T$.
In practice, we regress to the mean and diagonal covariance of a multi-dimensional
Gaussian density.
We implemented $p\left(\ki=1|\T\right)$ and $p\left(\ki=1|\Ci\right)$,
discriminators for binary variable $\ki$ conditioned on $\T$ and $\Ci$ respectively,
with regressors from $\T$ and $\Ci$ followed by sigmoid to normalize the output value to be in $[0,1]$.
We refer to the prior flow $p(\T|\Z)$ as the $\T$-flow, and the flows in each conditional prior $p(\Z|\Ki)$
as a $\Z$-flow.

All experiments were executed on a single NVIDIA TITAN Xp GPU with 12GB, and
optimized with Pytorch ADAM optimizer \cite{Kingma2014}, with default parameters and $lr = 1e-5$,
a warm up scheduler \cite{Vaswani2017} $warmup\_steps = 4000$, and mini-batch size of 10.
Further details are included in the supplemental material.
\section{Experiments} \label{sec:experiments}

To demonstrate the versatility of \TZK we train on up to six image datasets
(Table \ref{tab:datasets}), in unsupervised and semi-supervised settings.
All images were resized to $32\!\times\!32$ as needed, MNIST images were centered and
padded to $32\!\times\!32$.  When using grayscale (GS) images in an RGB setting,
the GS channel was duplicated in R, G, and B channels.

In all experiments below the images, $\T$, and class labels, $\ki$, for different
tasks are given.  The latent codes, $\Ci$, are not. In this semi-supervised  context
we sample the missing $\Ci$ according to the model, $\Ci \sim \Mopt$.
Specifically, at every mini-batch, we randomly choose $\Menc$ or $\Mdec$ with
equal probability.  When $\Mdec$ is chosen we sample from $\pdec ( \Ci )$, and for
$\Menc$ we return the marginal over $\penc ( \Ci|\ki, \T )$ with
respect to the observed binary variable $\ki$.

We chose CIFAR10 and MNIST as targets for conditional model learning.
Each comprises just  3.2\% of the entire multi-data training set of 1,892,916 images.
Table \ref{tab:experiments-summary} gives performance benchmarks in terms of
negative log-likelihood in bits per dimension (NLL) for existing flow-based models.

\begin{table}[t]
\centering
\scalebox{0.7}{
\begin{tabular}{|l|l|l|l|l|}
\hline
\textbf{Dataset} & \textbf{\begin{tabular}[c]{@{}l@{}}Image\\ Format\end{tabular}} & \textbf{\begin{tabular}[c]{@{}l@{}} \# Images \\ train / val \end{tabular}} & \% & \textbf{Classes} \\ \hline \hline
CIFAR10 & $32\!\times\!32$ RGB  & 50,000 / 10,000 & 3.2  & 10 \\ \hline
MNIST  & $28\!\times\!28$ GS  & 60,000 / 10,000 & 3.2  & 10 \\ \hline
Omniglot \textdagger & $105\!\times\!105$ RGB & 19,280 / 13,180 & 1.7 & NA \\ \hline
SVHN \textdagger & $32\!\times\!32$ RGB & 73,257 / 26,032 & 5.3 & 10  \\ \hline
ImageNet \textdagger & Varying RGB & 1,281,167 / 150,000 & 75.8 & 1000 \\ \hline
Celeba \textdagger  & $178\!\times\!218$ RGB & 200,000 / NA & 10.8 & NA \\ \hline
\end{tabular}
}
\vspace*{-0.1cm}
\caption{\label{tab:datasets}
Datasets marked with \textdagger \, were used in unsupervised settings only.
GS denotes grayscale images.
The {\em multi-data} training set consists of all six datasets, namely,
CIFAR10 \cite{Krizhevsky2009a}, MNIST \cite{LeCun1998}, Omniglot \cite{Lake2015},
SVHN \cite{Netzer2011}, ImageNet \cite{Russakovsky2014}, Celeba \cite{Liu2015}.
There are 1,892,916 images in total.
\% gives each dataset's fraction of the entire multi-data training set.}
\end{table}

\begin{table}[t]
\vspace*{-0.1cm}
\centering
\scalebox{0.7}{
\begin{tabular}{|l|l|l|l|l|l|}
\hline
 & \textbf{Glow} & \textbf{FFJORD} & \textbf{RealNVP} & \textbf{TzK Prior} & \textbf{TzK Cond.}  \\ \hline \hline
CIFAR10 & 3.35 & 3.4    & 3.49 & 3.54 & \textbf{2.99} *    \\ \hline
MNIST & 1.05 \textdagger\textdagger & \textbf{0.99} & 1.06 \textdagger\textdagger & 1.11 &  1.02 * \textdagger\\ \hline
\end{tabular}
}
\vspace*{-0.1cm}
\caption{\label{tab:experiments-summary}
Comparison of negative log-likelihood bits per dimension (NLL) on test data (lower is better).
*Results of dataset conditional prior. \textdagger Model was trained on all 6 datasets (Table \ref{tab:datasets}).
We compare to Glow \cite{Kingma2018}, FFJORD \cite{Grathwohl2018}, and RealNVP \cite{Dinh2016a}.
Results marked with \,\textdagger\textdagger \, are taken from \cite{Grathwohl2018}.
\vspace*{-0.1cm}
}
\end{table}

All learning occurred in an online fashion, adding new conditional knowledge types as needed.
When training begins, we start with a model with no knowledge, \ie, $\Knum = 0$, which is
just a Glow probability density estimator. As data are sampled for learning, new knowledge
types are added only when observed, in a semi-supervised manner, \ie, the class label
is given, the latent code $\Ci$ is not. In most of the experiments below the only
class label used is the identity of the dataset from which the image was drawn.

\begin{figure}[t]
    \centering
    \begin{minipage}[b]{.5\columnwidth}
    \centering
    \includegraphics[width=0.97\textwidth]{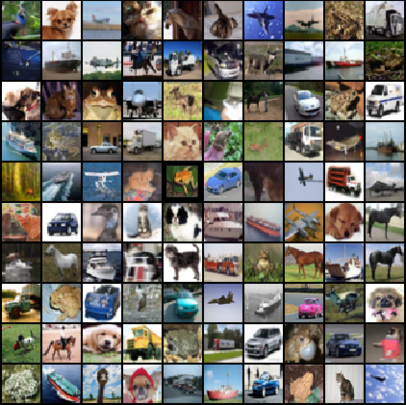}
        \end{minipage}%
            \begin{minipage}[b]{.5\columnwidth}
    \centering
    \includegraphics[width=0.97\textwidth]{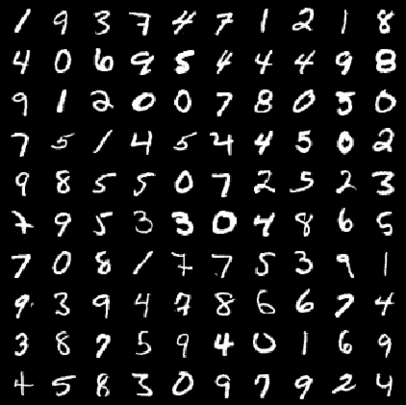}
        \end{minipage}
    \vspace*{-0.7cm}
    \caption{
    Random training samples from CIFAR10 and MNIST.
    }
    \label{fig:samples-data}
\end{figure}

\begin{figure}[t]
    \centering
    \begin{minipage}[b]{.5\columnwidth}
    \centering
    \includegraphics[width=0.97\textwidth]{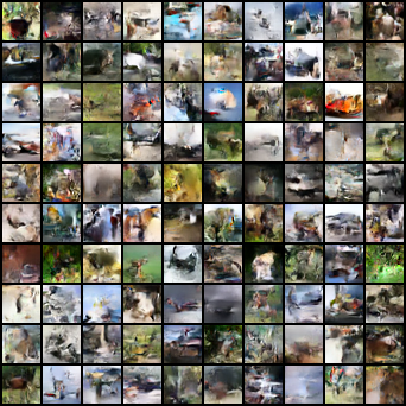}
    \subcaption{CIFAR10 only}\label{fig:cifar10-only-sample}
    \end{minipage}%
        \begin{minipage}[b]{.5\columnwidth}
    \centering
    \includegraphics[width=0.97\textwidth]{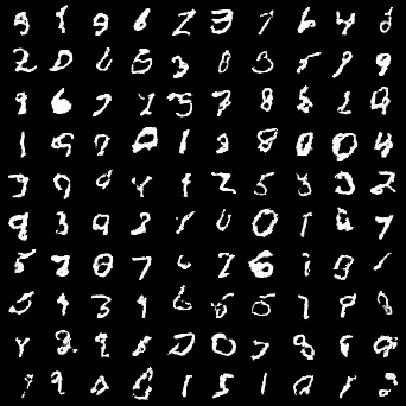}
    \subcaption{MNIST only}\label{fig:mnist-only-sample}
    \end{minipage}
    \vspace*{-0.7cm}
    \caption{
    Random samples from two baseline models, each trained with a single dataset (CIFAR10 and MNIST).
    The NLL for the CIFAR10 model is 3.54.  The NLL for the MNIST model is 1.11.
    }
    \label{fig:samples-priors-1-data}
\end{figure}

\begin{figure}[t]
    \centering
    \begin{minipage}[b]{.5\columnwidth}
    \centering
    \includegraphics[width=0.97\textwidth]{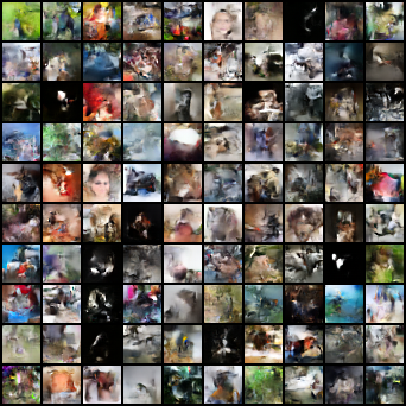}
    \subcaption{multi-data}\label{fig:multi-data-sample}
    \end{minipage}%
        \begin{minipage}[b]{.5\columnwidth}
    \centering
    \includegraphics[width=0.97\textwidth]{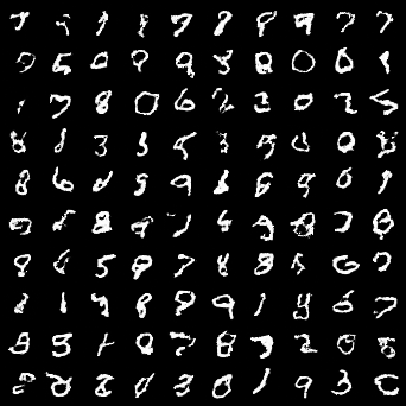}
    \subcaption{MNIST+Omniglot}\label{fig:multi-data-bw-sample}
    \end{minipage}
    \vspace*{-0.7cm}
    \caption{
    Samples from a model trained on all six datasets (\ref{fig:multi-data-sample}),
    and from one trained on MNIST+Omniglot (\ref{fig:multi-data-bw-sample}).
    Sample quality is similar to models trained solvely on CIFAR10 and MNIST (Fig.\  \ref{fig:samples-priors-1-data}), despite slightly higher NLL (3.6 for multi-data model
    and 1.28 for MNIST+Omniglot model). Samples are more diverse, however, reflecting the greater
    heterogeneity of the training data.
                }
    \label{fig:samples-priors-multi-data}
\end{figure}

\subsection{Baselines} \label{sec:baseline}

Two baseline models are trained on CIFAR10 and MNIST, training samples
for which are shown in Fig.\ \ref{fig:samples-data}.
Each used a Glow architecture for the $\T$-flow, with  512 channels,
32 steps, and 3 layers.  (See \cite{Kingma2018} for more details.)
These models give test NLL values of 3.54 and 1.11, comparable to the
state-of-the-art with flow-based models. Differences between our NLL numbers
and those reported for Glow by others in Table \ref{tab:experiments-summary}
are presumably due to implementation and optimization details.
Fig.\ \ref{fig:samples-priors-1-data} shows random samples from the two models,
the  quality of which compare well with training samples (Fig.\ \ref{fig:samples-data}).

When we train the same architecture on all 6 datasets (\ie, multi-data),
we obtain NLL of 3.6 when testing on CIFAR10.
Random samples from this model are shown in Fig.\ \ref{fig:multi-data-sample}.
One can clearly see the greater diversity of the training data, with images
resembling faces and grayscale characters for example.
When the same architecture is trained on the union of MNIST and Omniglot,
and tested on MNIST, the NLL is 1.28.
Random samples of this model (Fig.\ \ref{fig:multi-data-bw-sample})
again show greater diversity.
Although the NLL numbers with these models, both learned from larger
training sets, are slightly worse, the image quality remains similar to
models trained on a single dataset (Fig.\ \ref{fig:samples-priors-1-data}).

\begin{figure}[t]
    \centering
    \begin{minipage}[b]{.5\columnwidth}
    \centering
    \includegraphics[width=0.97\textwidth]{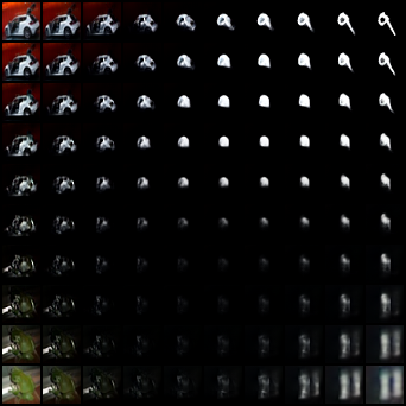}
    \subcaption{CIFAR10}
    \label{fig:interpolate-priors-cifar10}
    \end{minipage}%
        \begin{minipage}[b]{.5\columnwidth}
    \centering
    \includegraphics[width=0.97\textwidth]{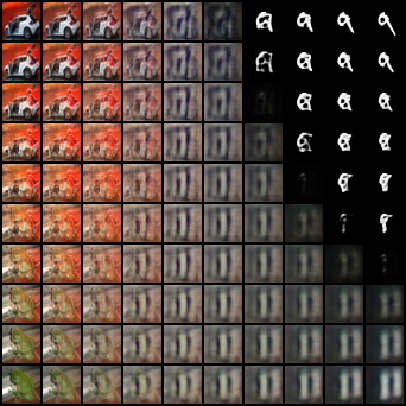}
    \subcaption{multi-data}
    \label{fig:interpolate-priors-multi-data}
    \end{minipage}
    \vspace*{-0.7cm}
    \caption{
    Given a model trained solely on CIFAR10, \ref{fig:interpolate-priors-cifar10}  depicts
    interpolation in $\Z$ between random samples from CIFAR10, MNIST, and SVHN.
    Interpolation reveals regions of $\Z$ that correspond to relatively poor quality images.
    This occurs even when the interpolated images are visually similar,
    and reflects relatively sparse coverage of the high-dimensional image space.
    Given a model trained on all six datasets (multi-data), the interpolation results
    in \ref{fig:interpolate-priors-multi-data} are much better than those above in \ref{fig:interpolate-priors-cifar10}.
    With more training data we obtain a denser model with visually better interpolation.
    }
\end{figure}

\subsection{Interpolation - Visualizing Flow Expressiveness}

Insight into the nature of the generative model can be gleaned
from latent space interpolation.  Here, given four images (observations $\T$),
we obtain latent space coordinates, $\Z = \fT^{-1}(\T)$.
We then linearly interpolate in $\Z$ before mapping back to $\T$ for visualization.
In a flow-based generative model with a Gaussian prior on $\Z$, we expect
interpolated points to have probability density as high or higher than
the end points, and at least as high as one of the two endpoints.

Despite Glow being a powerful model, the results in Fig.\ \ref{fig:interpolate-priors-cifar10}
reveal deficiencies. Training on CIFAR10 data produces a model that yields
interpolated images that are not always characteristic of CIFAR10
(e.g., the darkened images in Fig.\ \ref{fig:interpolate-priors-cifar10} ).
Even with color images (\ie, SVHN), which are expected to be represented reasonably
well by a CIFAR10 model, there are regions of low quality interpolants.

One would suspect that a model trained on the entire multi-data training set,
rather than just CIFAR10, would yield a better probability flow, exhibiting
denser coverage of image space. Consistent with this, Fig.\
\ref{fig:interpolate-priors-multi-data} shows superior interpolation
in $\Z$.

\subsection{Specializing a $\T$-Flow} \label{sec:specialization}

In this section we further explore the benefits of unsupervised training over large
heterogeneous datasets and the use of \TZK for learning conditional models in an online
manner. To that end, we assume a $\T$-flow has been learned and then remains fixed
while we learn one or more conditional models, as one might with unknown downstream tasks.
The Glow-like architecture used for the $\T$-flow (\ie, for $p(\T | \Z)$) had 512 channels,
20 steps, and 3 layers, a weaker model than those in \cite{Kingma2018} and the baseline
models above with 3 layers of 32 steps. The architecture used for the $\Z$-flow, for each of the
conditional models (\ie, for $p(\Z | \Ki)$), had one layer with just 4 steps.

In the first experiment the $\T$-flow is trained solely on CIFAR10 data,
entirely unsupervised.  The $\T$-flow was then frozen, and conditional models
were learned, one for CIFAR10 and one for MNIST.
Doing so exploits just one bit of supervisory information, namely, whether each
training image originated from CIFAR10 or MNIST.
Although this is a relatively weak form of supervision, the benefits are significant.
The MNIST images serve as negative samples for the conditional CIFAR10 model,
and {\em vice versa}. This allows the discriminators of the respective conditional
models to learn tight conditional distributions.

\begin{figure}[t]
    \centering
    \begin{minipage}[b]{.5\columnwidth}
    \centering
    \includegraphics[width=0.97\textwidth]{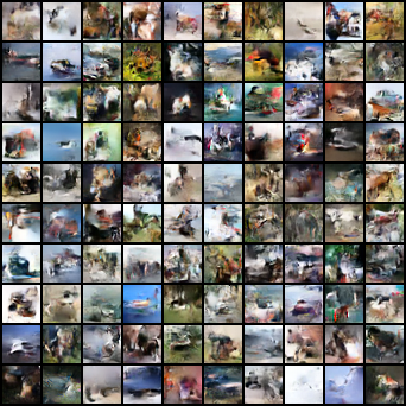}
    \subcaption{CIFAR10 conditional}\label{fig:cifar10-cond-cifar10-sample}
    \end{minipage}%
        \begin{minipage}[b]{.5\columnwidth}
    \centering
    \includegraphics[width=0.97\textwidth]{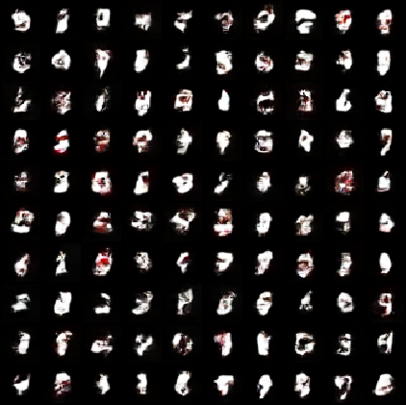}
    \subcaption{MNIST conditional}\label{fig:cifar10-cond-mnist-sample}
    \end{minipage}
    \vspace*{-0.7cm}
    \caption{
    The ability of \TZK to learn tight conditional priors is demonstrated here by freezing a
    $\T$-flow trained on CIFAR10 only, then learning conditional priors using CIFAR10 and MNIST.
    Random samples from the CIFAR10 conditional are shown in \ref{fig:cifar10-cond-cifar10-sample}.
    When tested on CIFAR10, the NLL for this model is just 2.99.
    Random samples from the MNIST conditional, in \ref{fig:cifar10-cond-mnist-sample},
    are surprisingly good given that MNIST data was not used to learn the $\T$ flow.
    The NLL for the MNIST conditional, tested on MNIST, is 1.33.
    }
    \label{fig:sample-conditional-prior-cifar10}
\end{figure}

Indeed, the resulting CIFAR10 conditional model exhibits a significant performance
gain, with a NLL of 2.99 when evaluated on the CIFAR10 test set, at or better than
state-of-the-art for CIFAR10, and a great improvement over the baseline $\T$-flow
(with 20 steps per layer), the NLL for which was 3.71 on the same test set.
Fig. \ref{fig:cifar10-cond-cifar10-sample} shows random samples from the conditional model.

\begin{figure}[t]
    \centering
    \begin{minipage}[b]{.5\columnwidth}
    \centering
    \includegraphics[width=0.97\textwidth]{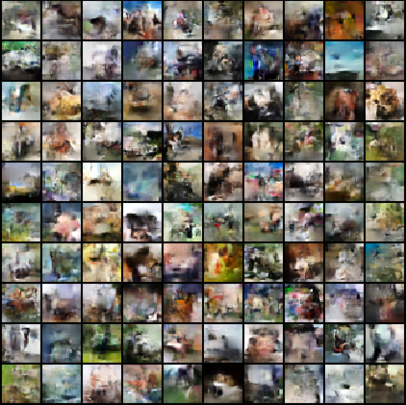}
    \subcaption{CIFAR10 conditional}\label{fig:multi-data-cond-cifar10-sample}
    \end{minipage}%
        \begin{minipage}[b]{.5\columnwidth}
    \centering
    \includegraphics[width=0.97\textwidth]{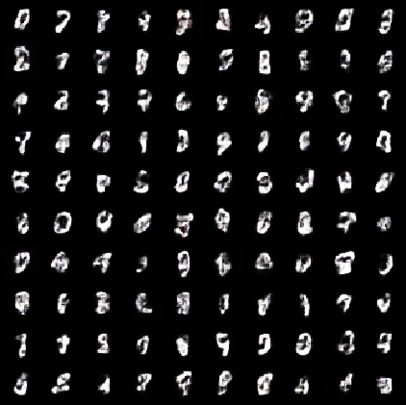}
    \subcaption{MNIST conditional}\label{fig:multi-data-cond-mnist-sample}
    \end{minipage}
    \vspace*{-0.7cm}
    \caption{
    \TZK offers a powerful framework to specialize a generative flow model trained in an
    unsupervised fashion on a large heterogeneous dataset.
    By learning tight conditional priors, these models are comparable to those trained end-to-end
    on a single dataset. Here, we train two conditional priors concurrently.
    Although trained concurrently, samples share the same latent representation $\Z$.
    The NLL for CIFAR10 (\ref{fig:multi-data-cond-cifar10-sample}) is 3.1.
    The NLL for MNIST (\ref{fig:multi-data-cond-mnist-sample}) is 1.02.
    }
    \label{fig:sample-conditional-prior-multi-data}
\end{figure}

Just as surprising is the performance of the MNIST conditional, even though the CIFAR10
data on which the $\T$-flow was trained did not contain images resembling the grayscale
data of MNIST.  Despite this, the conditional model was able to isolate regions of
the latent space representing predominantly grayscale MNIST-like images,
random samples of which are shown in Fig.\ \ref{fig:cifar10-cond-mnist-sample}.
When evaluated on MNIST data, the conditional model produced a NLL 1.33.
While these results are impressive, one would not expect a flow trained on
CIFAR10 to provide a good latent representation for many different image domains,
like MNIST.

In the next experiment we train a much richer $\T$-flow from
the entire multi-data training set of 1,892,916 images, again unsupervised.
Once frozen, we again learn conditional models for CIFAR10 and MNIST.
Despite MNIST and Omniglot representing a small fraction of the training set,
the MNIST conditional model exhibits state-of-the-art performance, with NLL
of 1.02 on the MNIST test set.  Random samples of the model are shown in
Fig.\ \ref{fig:multi-data-cond-mnist-sample}.
Similarly, the CIFAR10 conditional model exhibits state-of-the-art performance,
with NLL 3.1. While slightly worse than the model trained from CIFAR10,
it is still much better than our benchmark $\T$-flow, with 3
layers of 32 steps, and NLL of 3.54. Random samples from this CIFAR10
conditional model are shown in Fig.\ \ref{fig:multi-data-cond-cifar10-sample}.

In terms of cost, the time required to train the conditional models is
roughly half the time needed to train our baseline $\T$-flow model (or
equivalently Glow).
Freezing the $\T$-flow allows for asynchronous optimization of all conditional priors,
resulting in significant gains in training time, while still maintaining a model of
the joint probability. That is, conditional models can be trained in
parallel so the training does not scale with the number of knowledge types.
End-to-end training also benefits from this parallelism, but
does require synchronization for the shared $\T$-flow.
Finally, training a weaker $\T$-flow with 20 steps and 3 layers is marginally
faster than training a more expressive flow with 32 steps.

\subsection{End-to-End Hierarchical training}

We next consider a hierarchical extension to \TZK for learning larger models.
Suppose, for example, one wanted a \TZK model with 10 conditional priors, one
for each MNIST digit.
Conditioning on 10 classes in \TZK would require 10 independent discriminators,
and 40 independent regressors (2 for $\Ci$, 2 for $\Z$, per knowledge type $i$).
This does not scale well to large numbers of conditional priors.

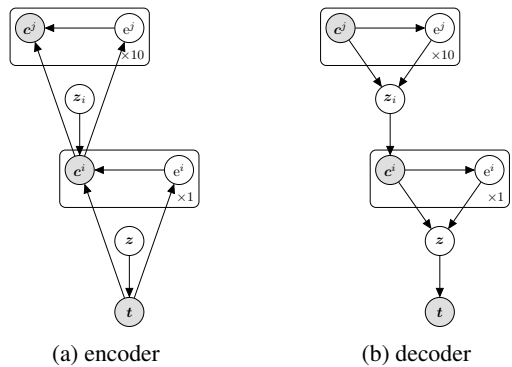
\begin{figure}[t]
    \centering
    \begin{minipage}[b]{.5\columnwidth}
    \centering \scalebox{0.55}{\begin{tikzpicture}

    \node[obs] (T) {$\T$};
  \node[latent, above=of T]                (Z) {$\Z$};
  \node[latent, above=of Z, xshift=1.2cm] (ki) {$\ki$};
  \node[obs, above=of Z, xshift=-1.2cm]     (Ci) {$\Ci$};

  \node[latent, above=of Ci]                (Zi) {$\Z_{i}$};
  \node[latent, above=of Zi, xshift=1.2cm] (kj) {$\kj$};
  \node[obs, above=of Zi, xshift=-1.2cm]     (Cj) {$\Cj$};

  \edge{Z} {T} ; \edge {T} {ki,Ci} ; \edge {ki} {Ci} ;
\edge{Zi} {Ci} ; \edge {Ci} {kj,Cj} ; \edge {kj} {Cj} ;

\plate {Ki} {(Ci)(ki)} {$\times 1$} ;
\plate {Kj} {(Cj)(kj)} {$\times 10$} ;

\end{tikzpicture}
}
    \subcaption{encoder}\label{fig:tzk-model-encoder-hier}
    \end{minipage}%
        \begin{minipage}[b]{.5\columnwidth}
    \centering \scalebox{0.55}{\begin{tikzpicture}

    \node[obs]                               (T) {$\T$};
  \node[latent, above=of T]                (Z) {$\Z$};
  \node[latent, above=of Z, xshift=1.2cm] (ki) {$\ki$};
  \node[obs, above=of Z, xshift=-1.2cm]     (Ci) {$\Ci$};

  \node[latent, above=of Ci]                (Zi) {$\Z_{i}$};
  \node[latent, above=of Zi, xshift=1.2cm] (kj) {$\kj$};
  \node[obs, above=of Zi, xshift=-1.2cm]     (Cj) {$\Cj$};

  \edge{Z} {T} ; \edge {ki,Ci} {Z} ; \edge {Ci} {ki} ;
\edge{Zi} {Ci} ; \edge {kj,Cj} {Zi} ; \edge {Cj} {kj} ;

\plate {Ki} {(Ci)(ki)} {$\times 1$} ;
\plate {Kj} {(Cj)(kj)} {$\times 10$} ;

\end{tikzpicture}
}
    \subcaption{decoder}\label{fig:tzk-model-decoder-hier}
    \end{minipage}
    \vspace*{-0.6cm}
    \caption{
    The modular nature of \TZK allows to build hierarchical model, where $\Ci$ of one \TZK model
    is serving as $\T$ of a domain-specific \TZK model. The joint objective is simple summation of the multiple objectives.
    This results in a model that supports fine grain control through sub-division of likelihood manifolds.
    }
    \label{fig:hierarchical-tzk}
\end{figure}

As an alternative, one can compose \TZK hierarchically.
For example, the first \TZK model could learn a conditional prior for MNIST images in general,
while the second model provides further specialization to digit-specific priors.
In particular, as depicted in Fig.\ \ref{fig:hierarchical-tzk}, the second \TZK model
takes as input observations the latent codes from the MNIST conditional model, and
then learns a second \TZK model comprising a new latent space,
on which the 10 digit-specific priors are learned.
The key advantage of this hierarchical \TZK model is that the latent code space
for the generic MNIST prior in the first \TZK model is low-dimensional, so
training the second \TZK model with 10 conditional priors is much more efficient
in terms of both training time and memory.

\begin{figure}[t]
    \centering
    \begin{minipage}[b]{.5\columnwidth}
    \centering
    \includegraphics[width=0.97\textwidth]{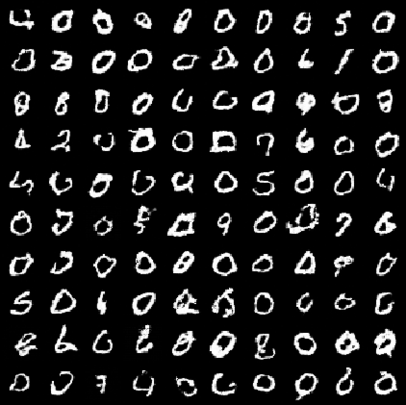}
    \subcaption{MNIST "0"}\label{fig:multi-data-bw-cond-mnist0-sample}
    \end{minipage}%
        \begin{minipage}[b]{.5\columnwidth}
    \centering
    \includegraphics[width=0.97\textwidth]{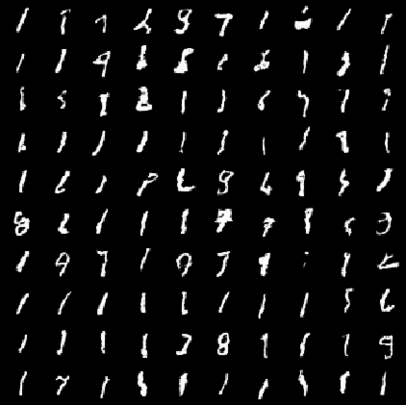}
    \subcaption{MNIST "1"}\label{fig:multi-data-bw-cond-mnist1-sample}
    \end{minipage}
    \vspace*{-0.6cm}
    \caption{
    A $\T$-flow was first learned with all datasets, and then frozen,
    after which an MNIST conditional over $\T$ was learned (the task) jointly with MNIST
    digit conditional over $\Ci$ (the task domain).
    This procedure allows efficient joint training for 10 conditional priors
    over a low-dimensional $\Ci \in \mathbb{R}^{\Cdim}$, as depicted in Fig.\ \ref{fig:hierarchical-tzk}.
    The NLL for the MNIST conditional prior (over $\T$) is 1.17.
    The NLL for the digit class conditional prior (over $\Ci$) is 1.06.
    }
    \label{fig:hierarchical-class-conditional-tzk}
\end{figure}

To implement this idea, the $\T$-flow in the first \TZK model comprised
3 layers with 10 steps, and 512 channels, a weaker flow than those used
in previous experiments with 20 or 32 steps. The $\Z$-flow for the MNIST conditional
model consisted of 512 channels and 1 layer of 4 steps, with a 10D latent
code space for $\Ci$.
The second stage \TZK model then maps the 10D $\Ci$ vectors to a latent
space with a probability flow comprising 1 layer of 4 steps with 64 channels,
on which 10 conditional priors are learned, each with a $\Z$-flow comprising one
more layer of 4 steps and 64 channels, with 2D latent codes $\Cj$.
Fig.\ \ref{fig:hierarchical-tzk} depicts the model.

Training this two-stage \TZK model entails optimization of the sum of two \TZK
losses, defined in Eq.\ \eqref{eq:tzk-entropy-loss-lower-bound}.
The first $\T$-flow was learned from the multi-data traing set, and then fixed, as this flow is
somewhat expensive to train. Everything else (\ie, the $\Z$-flow in the first \TZK model,
and all components of the second \TZK model) was trained jointly end-to-end.
The trained model had NLL 1.17 for the MNIST prior (first \TZK model),
and 1.06 for digit-specific priors; class-conditional samples are shown in
Fig.\ \ref{fig:hierarchical-class-conditional-tzk}.

This experiment also provides empirical evidence of consistency of
the encoder and decoder within a \TZK model.
Quantitatively, the digit-conditional discriminative model had 0.87 classification
accuracy over $\Ci$. Although far from state-of-the-art in classification accuracy,
we note that we allowed more than one class to high probability, rather than choosing a
single category with highest probability.
Instead, the model learned a joint distribution over 10 independent classes.
Qualitatively, the consistency can be observed in the samples in Fig.\ \ref{fig:hierarchical-class-conditional-tzk},
which are strongly correlated with the classification accuracy. In other words,
samples are roughly aligned with 0.87 classification accuracy. In effect, \TZK dual structure
results in consistency between the generative and discriminative components.
More importantly, this allows for multiple evaluation criteria of a \TZK generative model (\ie,
per conditional prior), in addition to NLL, which is not necessarily a meaningful quantity
to measure the performance of a generative model.

We perform one additional experiment with CIFAR10 conditional. This time the model predicted
a binary representation of the label (\ie, label 3 = $0011$) with 4 bits.
The trained model had NLL of 3.64 over the CIFAR10 conditional, and 0.74
classification of the domain task model.
The accuracy was independent per bit acknowledging similarity between classes,
as a result of the arbitrary division of classes according the label binary
representation. This last experiment again demonstrates the ability for \TZK
to learn compositional structure that represents a joint distribution.

\section{Conclusions} \label{sec:conclusion}

This paper introduces a versatile conditional generative model based on
probability flows.
It supports compositionality without a priori
knowledge of the number of classes or the relationships between classes.
Trained with maximum likelihood, it provides efficient inference and
sampling from class-conditionals or the joint distribution.
This allows one to train generative models from multiple
heterogeneous datasets, while retaining strong prior models over
subsets of the data (e.g., from a single dataset, class label,
or attribute).

The resulting model is efficient to train, either end-to-end,
in two phases (unsupervised flow followed by conditional models),
or hierarchically.
In addition, \TZK offers an alternative motivation for the use of MI in ML
models, as a natural term that arises given the assumption that the joint
distributions over observation and multiple latent codes has two equally
plausibly factorization of encoder and decoder.
Our experiments focus on models learned from six different image datasets,
with a relatively weak Glow architecture, conditioning on various types of
knowledge, including the identity of the source dataset, or class labels.
This yields log likelihood comparable to state-of-the-art, with compelling
samples from conditional priors.

\vspace*{-0.1cm}
\section*{Acknowledgements}

\vspace*{-0.15cm}
We thank Ethan Fetaya,  James Lucas, Alireza Makhzani, Leonid Sigal, and Kevin Swersky
for helpful comments on this work.
We also thank the Canadian Institute for Advanced Research
and NSERC Canada for financial support.

\bibliography{paper}
\bibliographystyle{icml2019}

\savestatus

\appendix
\clearpage
\section{Formulation Details}
\label{sec:tzk-detailed-formulation}

\subsection{Encoder and Decoder Consistency}

Here we discuss in greater detail how the learning algorithm encourages consistency
between the encoder and decoder of a \TZK model, beyond the fact that they are
fit to the same data, and have a consistent factorization
(see Eqs.\ \eqref{eq:tzk-encoder-compositionality} - \eqref{eq:tzk-decoder-K}).
To this end we expand on several properties of the model and the optimization procedure.

One important property of the optimization follows directly from the difference
between the gradient of the lower bound in Eq.\ (\ref{eq:tzk-entropy-loss-lower-bound})
and the gradient of the cross-entropy objective.
By moving gradient operator into the expectation using reparametrization, one
can express the gradient of lower bound in terms of the gradient of the
$\log \Mopt$ and the regularization term in Eq.\ (\ref{eq:tzk-entropy-loss-regularized}).
That is, with some manipulation one obtains
\begin{eqnarray}
\frac{\partial}{\partial\params} \left( \frac{\log\Menc+\log\Mdec}{2} \right) \, = \,
\frac{\partial}{\partial\params}\log\left(\! \frac{\Menc+\Mdec}{2}\! \right) \nonumber \\
+ \frac{1}{2}\frac{\left(\frac{\pdec}{\penc}-1\right)\frac{\partial}{\partial\params}\Menc+\left(\frac{\penc}{\pdec}-1\right)\frac{\partial}{\partial\params}\Mdec}{\Menc+\Mdec} \, .~
\label{eq:tzk-gradient-correction}
\end{eqnarray}
Consistent with the regularization term in Eq.\ (\ref{eq:tzk-entropy-loss-regularized}),
this shows that for any data point where a gap $\Menc > \Mdec$ exists, the gradient applied
to $\Mdec$ grows with the gap, while placing correspondingly less weight on the gradient applied
to $\Menc$.
The opposite is true when $\Menc < \Mdec$.
In both case this behaviour encourages consistency between the encoder and decoder.
Empirically, we find that the encoder and decoder become reasonably consistent
very early in the optimization process.

\paragraph{Numerical Stability:}

Instead of using a lower bound, one might consider direct optimization of Eq.\ \eqref{eq:tzk-entropy-loss}.
To do so, one must convert $\log \penc$ and $\log \pdec$ to $\penc$ and $\pdec$.
Unfortunately, this is likely to produce numerical errors,
especially with 32-bit floating-point precision on GPUs.
While various tricks can reduce numerical instability, we find that using the
lower bound eliminates the problem while providing the additional benefits outlined above.

\paragraph{Tractability:}

As mentioned earlier, there may be several ways to combine the encoder and decoder into
a single probability model.
One possibility we considered, as an alternative to $\Mopt$
in Eq. \eqref{eq:tzk-encoder-decoder}, is
\begin{equation}
\Mmodel = \frac{1}{\beta} \left( \Menc \,\Mdec \right)^{\frac{1}{2}} ~,
\label{eq:tzk-encoder-decoder-alt}
\end{equation}
where $\beta = \int ( \Menc \,\Mdec )^{\frac{1}{2}} d\T\, d\K$
is the partition function.
One could then define the objective to be the cross-entropy as above with a
regularizer to encourage $\beta$ to be close to 1, and hence consistency between
the encoder and decoder.
This, however, requires a good approximation to the partition function.
Our choice of $\Mopt$ avoids the need for a good value approximation by
using reparameterization, which results in unbiased low-variance gradient,
independent of the accuracy of the approximation of the loss value.

\subsection{\TZK Entropy and Mutual Information}

This section provides some context and a derivation for Eq.\ \eqref{eq:tzk-entropy}.

As discussed in Sec.\ \ref{sec:tzk-learning}, probability density normalizing flows
allow for efficient learning of arbitrary distributions by learning a mapping from
independent components to a joint target distribution \cite{Dinh2014}.
It follows that for a sufficiently expressive $\T$-flow the \TZK model factorization
assumed in \eqref{eq:tzk-encoder-compositionality} - \eqref{eq:tzk-decoder-K} does not
pose a fundamental limitation when learning joint, conditional distributions.
In other words, it is likely that there exist distribution flows with which $\pjoint$
can be factored according to the encoder and decoder factorizations in \TZK.

To that end, we can assume that with a sufficiently expressive model one can assume
that the dual encoder/decoder model can fit to the true data distribution reasonably well.
In the ideal case, where the encoder and decoder are consistent and equal to the
underlying data distribution, \ie\ $\pjoint = \penc = \pdec$, we obtain the following
result, which relates the entropy of the data distribution to the mutual information
between the data distribution and the latent space representation:
{\abovedisplayskip=5pt\begin{align}
& -H \left(\K,\T\right) & = \nonumber \\
& \E{\K,\T\sim\pjoint}{\log\pjoint\left(\K,\T\right)} & = \nonumber \\
& \E{\K,\T\sim\pjoint}{\log \left(\frac{1}{2}\penc\left(\K,\T\right) +\frac{1}{2}\pdec\left(\K,\T\right)  \right) } & = \nonumber\\
& \E{\K,\T\sim\pjoint}{\frac{1}{2}\log \penc \left(\K,\T\right)+\frac{1}{2} \log\pdec \left(\K,\T\right)} & = \nonumber\\
& \E{\K,\T\sim\pjoint}{
\log p\left(\T\right)+ \!\!
\begin{array}{l}
\!\! \frac{1}{2}\sum_{i}\left(\begin{array}{l}
\!\!
\log p\left(\Ki|\T\right)+\\
\!\!
\log p\left(\Ki\right)+\\
\!\!
\log p\left(\Z\left(\T\right)|\Ki\right)- \!\!\! \\
\!\!
\log p\left(\Z\left(\T\right) \right)
\end{array}\right)
\!\!\!\!
\end{array}} & = \nonumber \\
& -H\left(\T\right)+\frac{1}{2}\sum_{i}\left(\begin{array}{l}
H\left(\Z\right)-\\
\E{\Ki\sim \pjoint}{H\left(\Z|\Ki\right)}-\\
H\left(\Ki\right)-\\
\E{\T\sim \pjoint}{H\left(\Ki|\T\right)}
\end{array}\right) & = \nonumber \\
& -H\left(\T\right)+\frac{1}{2}\sum_{i}\left(\begin{array}{l}
I\left(\Z;\Ki\right) - H\left(\Ki\right) -\\
\E{\T\sim \pjoint}{H\left(\Ki|\T\right)}
\end{array}\right) & = \nonumber\\
& -H\left(\T\right)-\!\sum_{i}\!H\left(\Ki\right)+\frac{1}{2}\!\sum_{i} ( I(\Ki;\T)\!+\!I(\Z;\Ki))
\label{eq:tzk-entropy-derivation}
\end{align}
}
Eq.\ (\ref{eq:tzk-entropy-derivation}) illustrates an interesting connection
between ML and MI, assuming \TZK to be the true underlying model.
One can interpret ML learning of $\Mopt$ as a lower bound for the sum of the negative entropy
of observations $\T$, the negative entropy of latent codes $\K$, and the MI between the
observations $\T$ and the latent codes $\Ki$, and between the latent state $\Z$ and
the latent codes $\K$.
This formulation arises naturally from the \TZK representation of the data distribution,
as opposed to several existing models that use MI as a regularizer \cite{Hjelm2018,Chen2016,Dupont2018,Klys2018}.

An important property of the \TZK formulation is the lack of variational approximations
where an auxiliary distribution $q(z|x)$ is used to approximate $p(z|x)$.
As a consequence, it is hoped that a more expressive $\Mopt$ will be better
able to approximate  $\pjoint$, leading to a tighter lower bound; since, compared to
variational inference (VI), a more expressive $q$ does not necessarily guarantees a tighter
lower bound as it is restricted to tractable families. In addition, since  $\DKL{q}{p}$ is unknown,
VI does not offer a method to measure that gap.

\section{Architecture Details}
\label{sec:implementation-details}

The components of a \TZK model, \ie, the factors in Eqs.\ \eqref{eq:tzk-encoder-compositionality}
- \eqref{eq:tzk-decoder-K}, have been implemented in terms of parameterized
deep networks.  In somewhat more detail, the prior over $\T$ was implemented with:
\begin{itemize}
\itemsep 0.05cm
\item $\fT(\Z): \Z \in \mathbb{R}^{\Tdim} \rightarrow \T \in \mathbb{R}^{\Tdim}$ is our Glow-based implementation \cite{Kingma2018}. Flow details are included with each experiment.
\item $\Menc(\T): \emptyset \rightarrow \T \in \mathbb{R}^{\Tdim}$ is parameterized
in terms of a probability normalizing flow $\fT(\Z)$ from a multivariate standard normal
distribution $p(\Z)$.
\end{itemize}
All density probability flow $\fT$ had 3 layers (multi-scale) and steps defined in each experiment,
with 512 channels for regressors in affine
coupling transforms \cite{Kingma2018}.

The priors over latent codes $\Ci$ were implemented with:
\begin{itemize}
\itemsep 0.05cm
\item $\Mdec (\Ci )$, for $ \Ci \in \mathbb{R}^{\Cdim}$, is parameterized in terms of a
probability normalizing flow $\fCi (\cdot ) : \mathbb{R}^{\Cdim} \rightarrow \mathbb{R}^{\Cdim}$
from a multivariate standard normal distribution.
\end{itemize}
For $\Ci \in \mathbb{R}^{\Cdim}$, we use a Glow architecture with 1 layer, 4  steps, and $10\Cdim$ channels, where $\Cdim=10$, unless specified otherwise in an experiment. $\Ci$ was shaped to have all dimensions in a single channel $\Cdim\!\times\!1\!\times\!1$.

The discriminators associated with different knowledge types $p(\ki=1|\cdot)$,
conditioned on observation $\T$ or a latent  code $\Ci$, were implemented with:
\begin{itemize}
\itemsep 0.05cm
\item $\Menc(\ki|\T ): \T \in \mathbb{R}^{\Tdim} \rightarrow \left[0, 1\right]$
\item $\Mdec (\ki|\Ci ): \Ci \in \mathbb{R}^{\Cdim} \rightarrow \left[0, 1\right] $
\end{itemize}
All discriminators from $\T$ and $\Ci$ had 3 layers of $3\!\times\!3$ convolution with $10\Cdim$ channels
(unless specified otherwise in experiment details), followed by linear mapping to the target dimensionality of 1,
and a sigmoid mapping to normalize the output to be $[0,1]$.

The conditional priors over $\T$ and $\Ci$ are modeled with regressors from the corresponding $\Ci$ and $\T$
to the distribution parameters (\eg, mean and variance for Gaussian), similar to VAE. More explicitly, we implemented the priors with regressors to the mean and diagonal covariance matrix of a Gaussian, as described below:
\begin{itemize}
\itemsep 0.05cm
\item $\Menc (\Ci|\ki, \T ): \T \in \mathbb{R}^{\Tdim} \rightarrow \left\{\mu,\sigma\right\} \in \mathbb{R}^{\Cdim} \times \mathbb{R}^{\Cdim}$ is composed of a regressor to the mean $\mu$
and diagonal covariance $\sigma$ of a Gaussian base distribution, and $\fCi (\cdot )$,
an invertible function that serves as a probability flow. The two components comprise
a single parametric representation of a generic probability distribution. We condition the
density on $\ki$ by learning two separate sets of weights, i.e., for $\ki \in \left\{0,1\right\}$.
The flow uses the same Glow architecture as  $\fT$, for which the details are given below.
All $\fCi$ had 4 flow steps, with dimensionality of $\Cdim = 10$, unless specified otherwise.
\item $\Mdec (\T|\ki, \Ci ): \Ci \in \mathbb{R}^{\Cdim} \rightarrow \left\{\mu,\sigma\right\} \in \mathbb{R}^{\Tdim} \times \mathbb{R}^{\Tdim}$
is composed of a regressor to the mean $\mu$ and diagonal covariance $\sigma$ of a Gaussian base distribution, and $\fT\left(\Z\right)$. We condition the probability density on $\ki$ by learning two separated set of weights for $\ki \in \left\{0,1\right\}$.
\end{itemize}
All Gaussian regressors $\Ci \rightarrow \Z$ were implemented with a linear mapping
$\mathbb{R}^{\Cdim} \rightarrow \mathbb{R}^{80\times4\times4}$, 3 layers of $3\times3$
convolution layers with 512 channels, and final layer with 192 channels, resulting in $\Z \in \mathbb{R}^{192\times4\times4}$.
All Gaussian regressors $\T \rightarrow \Ci$ were implemented with $3\times3$ convolutional
layer with 80 channel, followed by 3 layers of alternating squeeze \cite{Kingma2018}
and  $3\times3$ convolutional layers with 80 channels, followed by linear layer to $\Cdim$.
All regressors had ActNorm to initialize inputs to be mean-zero with unit variance,
and the weights of the last layer were initialized to 0, as in \cite{Kingma2018}.

\section{Model Evaluation}

Here we explain in detail how we evaluate the model for the experiments.
To that end, we consider the evaluation of the negative log likelihood of a data
sample under the model, and the process for drawing a random sample from the
model.

Given a set of test samples, the NLL is defined as the average negative log likelihood of
the individual samples (\ie, assuming IID samples).
Evaluating the NLL for $\Menc(\T)$ is straightforward in terms of the flow $\fT$
and the latent Gaussian prior $p(\Z)$.

To evaluate the NLL for a conditional distribution, given $\T$, we first draw a
random sample $\Ci \sim \Menc(\Ci|\ki=1, \T)$.
We then use that sample to build $\Mdec(\T|\ki=1, \Ci)$. with which We
evaluate the log probability of $\T$.

We next explain the procedure for sampling from a class conditional hierarchical model.
To sample from a the digit "1" over a domain $\Ci$ we first sample from
$\Cj \sim \Mdec(\Cj)$, followed by $\Ci \sim \Mdec (\Ci|\kj\!=\!1, \Cj)$, and
finally $\T \sim \Mdec(\T|\ki\!=\!1, \Ci)$.

\end{document}